\documentclass[oneside]{article}
\usepackage[english]{babel}
\usepackage[T1]{fontenc}
\usepackage{lmodern}
\usepackage[style=numeric, backend=biber, sorting=none]{biblatex}
\usepackage{amsfonts}
\usepackage{amsmath}
\usepackage{microtype}
\usepackage{graphicx}
\usepackage{authblk} 
\usepackage{subcaption}
\usepackage{csquotes}
\usepackage{booktabs}
\usepackage{abstract}
\usepackage[skip=0.5em, position=top]{caption}
\usepackage{geometry}
\usepackage{enumitem}
\usepackage{etoolbox}
\usepackage{hyperref}
\usepackage{etoolbox}          
\AfterEndEnvironment{section}{\par\noindent\ignorespaces}
\makeatletter
\patchcmd{\@enum@}{\@newctr}{\@newcounter}{}{}
\makeatother
\begin{document}

	\title{RF-BayesPhysNet: A Bayesian rPPG Uncertainty Estimation Method for Complex Scenarios}
\author[1,†]{Rufei Ma}
\author[1,*]{Chao Chen}
\affil[1]{Shandong Artificial Intelligence Institute, Qilu University of Technology (Shandong Academy of Sciences)}
\affil[*,†]{Corresponding: chench@qlu.edu.cn, aaaaidc@gmail.com}
\date{}

		\maketitle
	
	\begin{abstract}
		\setlength{\parindent}{0pt}
		\noindent
Remote photoplethysmography (rPPG) technology infers heart rate by capturing subtle color changes in facial skin using a camera, demonstrating great potential in non-contact heart rate measurement. However, measurement accuracy significantly decreases in complex scenarios such as lighting changes and head movements compared to ideal laboratory conditions. Existing deep learning models often neglect the quantification of measurement uncertainty, limiting their credibility in dynamic scenes. To address the issue of insufficient rPPG measurement reliability in complex scenarios, this paper introduces Bayesian neural networks to the rPPG field for the first time, proposing the Robust Fusion Bayesian Physiological Network (RF-BayesPhysNet), which can model both aleatoric and epistemic uncertainty. It leverages variational inference to balance accuracy and computational efficiency. Due to the current lack of uncertainty estimation metrics in the rPPG field, this paper also proposes a new set of methods, using Spearman correlation coefficient, prediction interval coverage, and confidence interval width, to measure the effectiveness of uncertainty estimation methods under different noise conditions. Experiments show that the model, with only double the parameters compared to traditional network models, achieves a MAE of 2.56 on the UBFC-RPPG dataset, surpassing most models. It demonstrates good uncertainty estimation capability in no-noise and low-noise conditions, providing prediction confidence and significantly enhancing robustness in real-world applications. We have open-sourced the code at \href{https://github.com/AIDC-rPPG/RF-Net}{https://github.com/AIDC-rPPG/RF-Net}
	\end{abstract}
	
	\section{Introduction}

Remote photoplethysmography (rPPG) is a non-contact physiological signal monitoring technology. It extracts physiological parameters such as heart rate from video by analyzing subtle color changes on the face or skin surface \cite{review}. Compared to traditional contact-based PPG, rPPG does not require direct skin contact, offering higher comfort and broad application prospects.

The basic principle of rPPG technology is based on the optical property fluctuations of the skin caused by blood volume changes. When the heart beats, blood circulation changes the way the skin absorbs and reflects light. Camera equipment can capture these subtle changes in color and brightness, and by processing video frame sequences, extract signals related to heart rate \cite{jbyl}. In recent years, rPPG technology has made significant progress in the fields of computer vision and signal processing, mainly in the following aspects:

\begin{enumerate}
	\item \textbf{Improvement of signal extraction algorithms}: Early methods were mostly based on color space conversion and signal processing, such as Poh et al. using independent component analysis (ICA) \cite{ICA} to extract heart rate signals. Subsequently, methods such as principal component analysis (PCA) \cite{PCA} and the plane orthogonal to skin (POS) \cite{POS} method were introduced to improve signal quality and robustness.
	
	\item \textbf{Application of deep learning}: With the rise of deep learning, researchers began using models such as convolutional neural networks (CNN) to directly learn heart rate signals from video \cite{firstdl}. These methods can automatically extract features and adapt to complex environmental changes.
	
	\item \textbf{Expansion of application areas}: The application of rPPG technology has expanded from medical health monitoring to emotion recognition, driver fatigue detection, human-computer interaction, and other fields, showing broad application prospects.
\end{enumerate}

However, rPPG technology still faces many challenges, such as:

\begin{itemize}
	\item \textbf{Motion artifacts}: Head or body movements of the subject interfere with the accurate extraction of heart rate signals.
	\item \textbf{Lighting variations}: Changes in environmental lighting affect the quality of images captured by the camera.
	\item \textbf{Skin type differences}: Different skin colors and characteristics affect the intensity of the reflected light signal.
\end{itemize}
To solve these problems, many solutions have been tried, such as eliminating noise through color space projection \cite{zs1}, \cite{zs2}, or certain skin reflection models \cite{pf1}, \cite{pf2}, to enhance the usability of the model in real-world scenarios. However, in practical applications, noise may have a completely different distribution from the ideal noise during training, leading to a decrease in model accuracy, and users cannot know the accuracy drop, making the model difficult to use. Based on the above analysis, rather than enhancing the model's robustness through denoising during training and still not knowing the model's accuracy in real scenarios, it is better to let the model output uncertainty, allowing users to decide whether to trust the model based on the uncertainty value.

Our main contributions are as follows:
\begin{itemize}
	\item Propose a novel neural network, the Robust Fusion Bayesian Physiological Network (RF-BayesPhysNet), for rPPG tasks.
	\item Propose a set of metrics for uncertainty estimation in the rPPG field.
	\item Introduce Bayesian neural networks
\end{itemize}

\section{Related Work}

\subsection{Traditional rPPG Signal Processing Methods}

Traditional rPPG signal processing methods primarily focus on extracting clean rPPG signals from facial videos containing noise. These methods are usually based on signal processing techniques, utilizing the characteristics of rPPG signals in specific frequency bands for extraction. Here are some commonly used traditional rPPG signal processing methods:

\begin{itemize}
	\item \textbf{Independent Component Analysis (ICA)}: 
	ICA \cite{ICA} is a blind source separation technique that assumes the observed signals are linear mixtures of multiple independent source signals. In rPPG signal extraction, ICA can be used to separate signal components related to cardiac activity, thereby removing the influence of other noise sources.
	
	\item \textbf{Principal Component Analysis (PCA)}: 
	PCA \cite{PCA} is a dimensionality reduction technique that represents the main features of data by finding the principal components with the largest variance. In rPPG applications, PCA can be used to extract the main components of rPPG signals, reducing the impact of noise.
\end{itemize}

These traditional methods can extract rPPG signals to some extent, but their performance is often limited in complex noise environments such as motion artifacts and lighting changes.

\subsection{Deep Learning-Based rPPG Methods}

With the rapid development of deep learning technology, deep learning-based rPPG methods have made significant progress in recent years, demonstrating greater robustness and accuracy compared to traditional signal processing methods. This section introduces several major deep learning-based rPPG methods. Deep learning-based rPPG methods can be divided into three categories: (1) CNN models dominated by spatial features (e.g., DeepPhys); (2) RNN/LSTM models for temporal modeling; (3) PhysNet and its variants for spatiotemporal joint modeling.

\subsubsection{Convolutional Neural Network (CNN) Methods}

Due to their excellent performance in image processing, convolutional neural networks have been widely applied to rPPG signal extraction. Chen and McDuff \cite{deepphys} proposed DeepPhys, an end-to-end two-stage spatiotemporal convolutional network that directly extracts pulse signals from facial video frames. The network first uses a motion representation network to capture subtle facial motion features, and then processes these features through an appearance representation network to extract pulse signals. The innovation of DeepPhys lies in introducing attention mechanisms and normalization, allowing the model to better adapt to different lighting conditions and skin types.

Wang et al. \cite{multask} proposed a multitask convolutional neural network that simultaneously performs rPPG signal extraction and facial region segmentation tasks. By sharing feature learning, this method enhances the model's perception of key facial regions, thereby improving the accuracy of pulse signal extraction.

\subsubsection{PhysNet and Its Variants}

PhysNet \cite{CNNLSTM} is an end-to-end deep learning model specifically designed for rPPG signal extraction, combining 3D convolution and temporal convolution to simultaneously process spatial and temporal information. The model uses temporal continuous waveform supervision signals instead of just predicting average heart rate, thereby better capturing heart rate variability information.

Botina-Monsalve et al. \cite{qlh} proposed RTrPPG, an improved version of PhysNet, which reduces model parameters and speeds up inference by adjusting input size and network depth, and improves accuracy by projecting to YUV color space.

Niu et al. \cite{Rhy} developed RhythmNet, a method that uses spatiotemporal convolutional neural networks to simultaneously extract heart rate and respiration rate from facial videos. This method constrains the network output in the frequency domain by designing specific loss functions, ensuring the main frequency of the predicted signal matches the reference signal.

\subsubsection{Challenges and Limitations}

Despite significant progress in deep learning-based rPPG methods, several challenges remain: (1) Limited generalization ability of models, with performance potentially declining in different environments, lighting conditions, and populations (e.g., different skin colors); (2) Most methods lack quantification of uncertainty in prediction results, making it difficult to assess the reliability of model predictions, thus challenging practical use.

These challenges are precisely what this paper aims to address with the Bayesian neural network-based rPPG method, particularly innovations in uncertainty estimation, which will be discussed in detail in the following sections.

\section{Model Architecture}
\subsection{Basics of Bayesian Neural Networks}

Bayesian Neural Networks (BNNs) are probabilistic extensions of traditional neural networks that express model uncertainty by replacing deterministic weights with probability distributions \cite{KL}. In BNNs, the network weights $\boldsymbol{w}$ are no longer fixed values but follow a prior distribution $p(\boldsymbol{w})$. Given training data $\mathcal{D}$, the goal of BNNs is to infer the posterior distribution of weights $p(\boldsymbol{w}|\mathcal{D})$, which can be computed using Bayes' theorem:

\begin{equation}
	p(\boldsymbol{w}|\mathcal{D}) = \frac{p(\mathcal{D}|\boldsymbol{w})p(\boldsymbol{w})}{p(\mathcal{D})}
\end{equation}

where $p(\mathcal{D}|\boldsymbol{w})$ is the likelihood function representing the probability of the data given the weights, and $p(\mathcal{D})$ is the marginal likelihood, which is typically difficult to compute directly. In practice, approximate methods such as variational inference are often used to estimate the posterior distribution \cite{bftd}.

Variational Inference (VI) offers an efficient approximation method. The core idea of variational inference is to introduce a parameterized family of probability distributions $\mathcal{Q}$, called the variational distribution family, and select an optimal distribution $q(w|\theta) \in \mathcal{Q}$ to approximate the true posterior distribution $p(w|D)$. Here, $\theta$ represents the variational parameters that control the shape of the variational distribution.

The goal of choosing the variational distribution $q(w|\theta)$ is to make it as close as possible to the true posterior distribution $p(w|D)$. Typically, we use the Kullback-Leibler (KL) divergence to measure the difference between the two distributions:

\begin{equation}
	\text{KL}(q(w|\theta) \,||\, p(w|D)) = \int q(w|\theta) \, \log \frac{q(w|\theta)}{p(w|D)} \, dw
\end{equation}

Since the KL divergence still contains the term $p(w|D)$ which is difficult to compute, directly minimizing the KL divergence is not feasible. Therefore, we maximize the Evidence Lower Bound (ELBO) instead:

\begin{equation}
	\text{ELBO}(q) = \mathbb{E}_{q(w|\theta)}[\log p(D|w)] - \text{KL}(q(w|\theta) \,||\, p(w))
\end{equation}
It can be proven that maximizing ELBO is equivalent to minimizing the KL divergence.

\subsection{RF-BayesPhysNet Architecture}
We propose a novel network architecture: Robust Fusion Bayesian Physiological Network (RF-BayesPhysNet) applied to Bayesian neural networks.

\begin{figure}[htbp]
	\centering
	\includegraphics[width=0.8\textwidth]{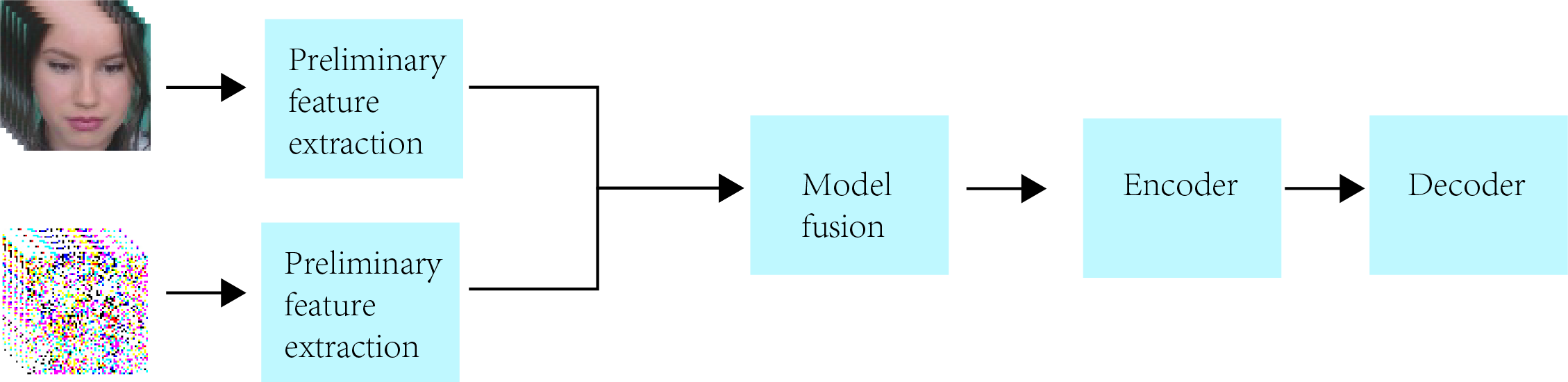}
	\caption{RF-BayesPhysNet Network Architecture}
	\label{fig:myimage}
\end{figure}

\begin{figure}[htbp]
	\centering
	
	\begin{subfigure}[t]{0.45\textwidth}
		\centering
		\includegraphics[width=\linewidth]{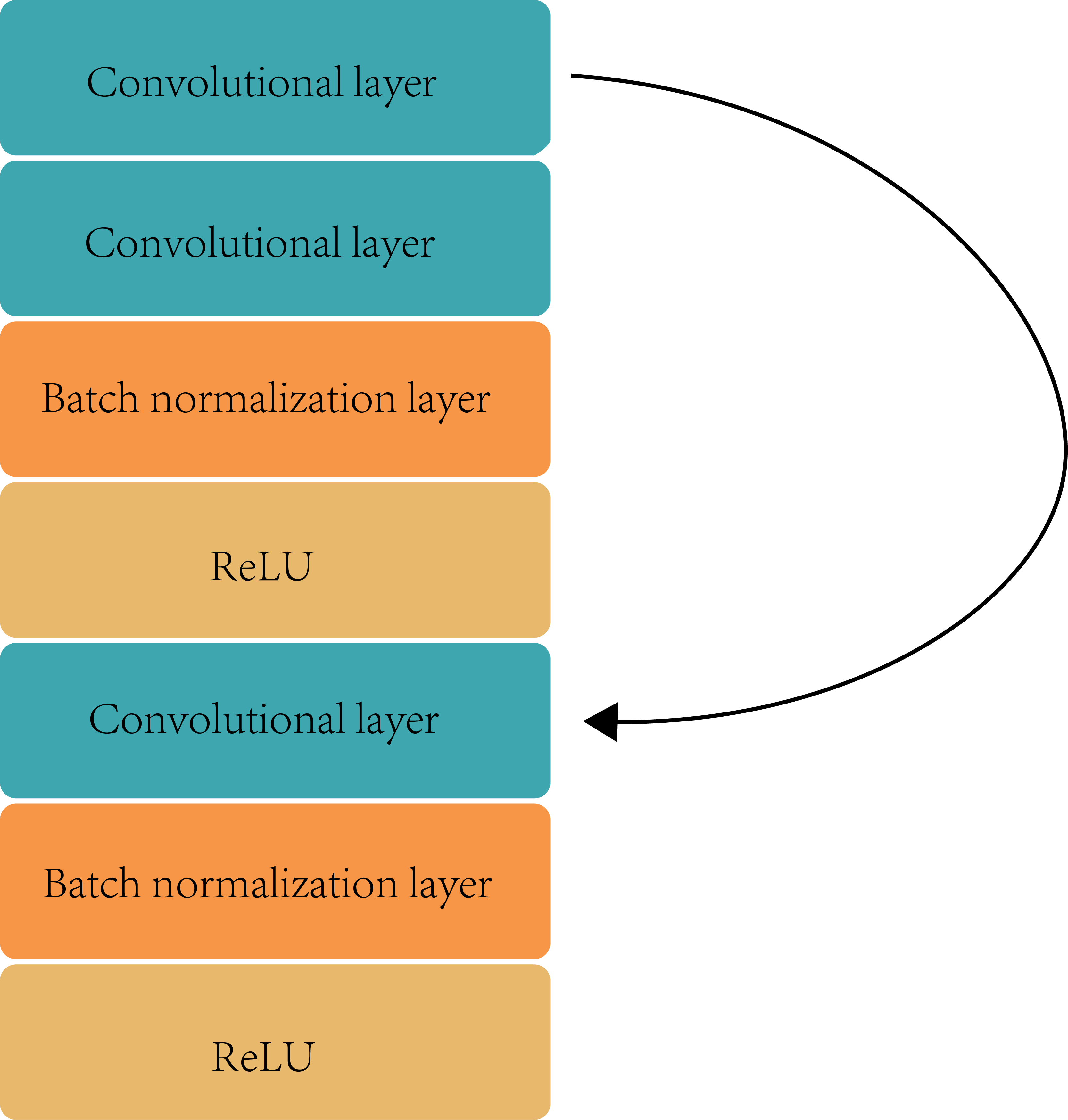}
		\caption{Differential Input Processing Branch Structure}
		\label{fig:subfig-a}
	\end{subfigure}%
	\hfill
	\begin{subfigure}[t]{0.45\textwidth}
		\centering
		\includegraphics[width=\linewidth]{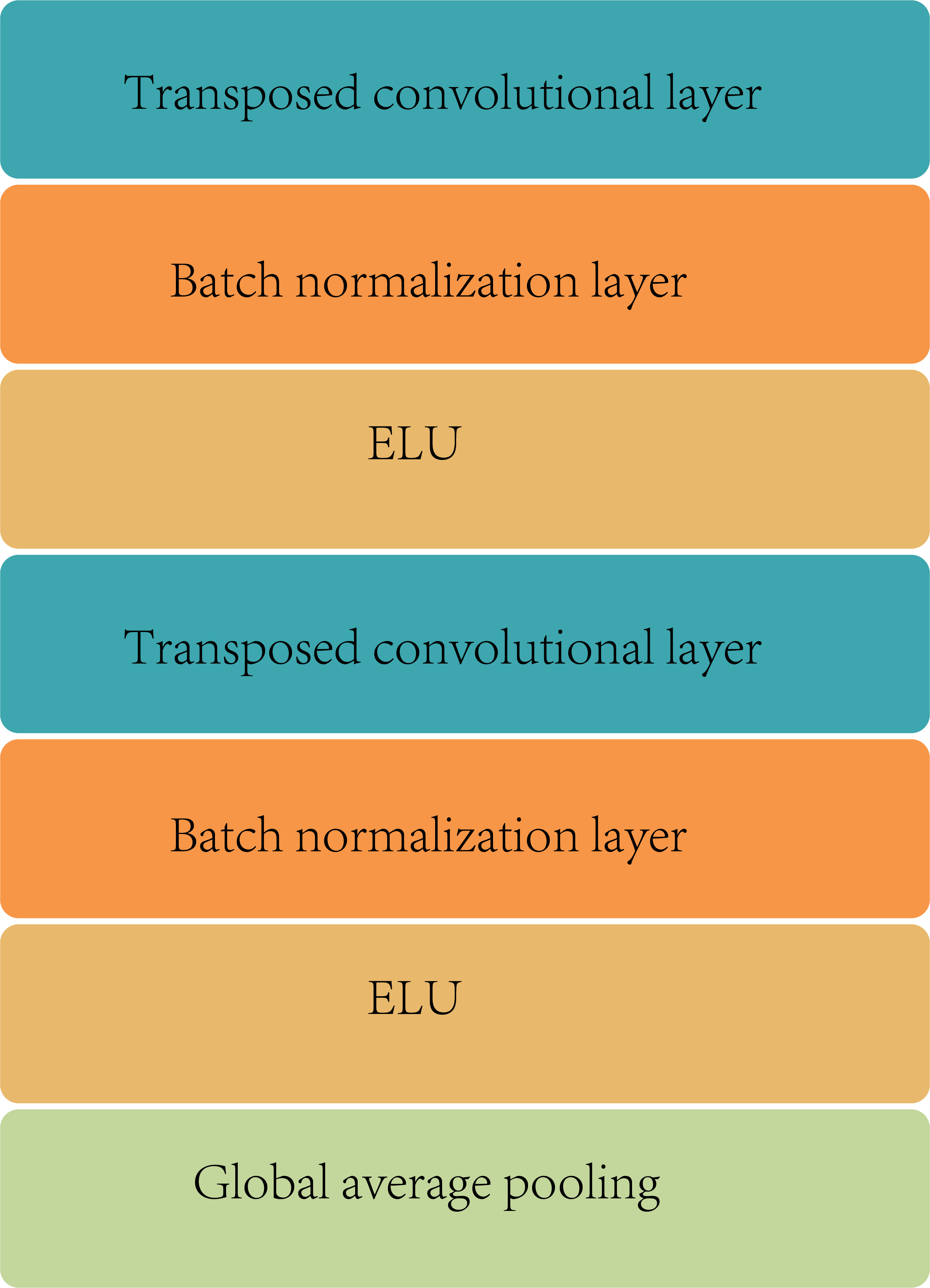}
		\caption{Decoder Structure}
		\label{fig:subfig-b}
	\end{subfigure}
	
	\caption{Overall Caption with Two Subfigures}
	\label{fig:two_images}
\end{figure}

Compared to fixed weights in traditional neural networks, our dynamic weight computation method is as follows:
Sample $\varepsilon$ from the standard normal distribution $\mathcal{N}(0, 1)$, and then compute the weights using equation (\ref{eq:reparam}):

\begin{equation}
	weight = \mu + \sigma \cdot \varepsilon  \label{eq:reparam}
\end{equation}
The RF-BayesPhysNet model designed in this study \ref{fig:myimage} is built on a U-Net structure and includes the following main components:

\begin{enumerate}
	\item \textbf{Initial Feature Extraction Module}: Processes the raw facial video sequence to capture spatial and temporal features. The module includes a 3D Bayesian convolution layer, a normalization layer, an activation function, and a max 3D pooling layer to ensure the size remains consistent for subsequent model fusion.
	\item \textbf{Differential Input Processing Branch}: The differential branch enhances the capture of motion artifacts by computing differences between adjacent frames, complementing the spatial features of the raw branch, thus improving the model's robustness to head movements. The module consists of 3D convolution layers, normalization layers, and activation functions, linked by residual methods as shown in \ref{fig:subfig-a}.
	\item \textbf{Feature Fusion Module}: Fuses the features from the raw and differential branches. The outputs from the initial feature extraction module and the feature fusion module are stacked to complete this process.
	\item \textbf{Encoder}: Gradually extracts higher-level feature representations. Composed of multiple 3D convolution layers, normalization layers, and activation functions, as shown in \ref{fig:subfig-a}.
	\item \textbf{Decoder}: Gradually restores the temporal resolution of the signal through deconvolution operations, ultimately outputting the heart rate signal, as shown in \ref{fig:subfig-b}.
\end{enumerate}

A key innovation of the model is its flexible Bayesian layer configuration. The model allows users to specify which layers in the network should be Bayesian layers and which should remain as traditional deterministic layers, providing a balance between computational complexity and uncertainty estimation capability.

\subsection{Overview of Uncertainty Estimation Methods}

In machine learning models, uncertainty is typically divided into two main types \cite{bqdx}:

\begin{enumerate}
	\item \textbf{Aleatoric Uncertainty}: Aleatoric uncertainty arises from noise and randomness inherent in the data and cannot be reduced by collecting more data. In the rPPG field, this may come from sensor noise, environmental light changes, or subject movement.
	
	\item \textbf{Epistemic Uncertainty}: Epistemic uncertainty arises from the model's incomplete knowledge of the world and can be reduced by increasing the amount of training data. It reflects the model's uncertainty about unseen data.
\end{enumerate}

In this study, we estimate the prediction uncertainty of RF-BayesPhysNet through the following methods:

\begin{equation}
	\hat{y} = \mathbb{E}_{p(\boldsymbol{w}|\mathcal{D})}[f_{\boldsymbol{w}}(\boldsymbol{x})] \approx \frac{1}{T}\sum_{t=1}^{T}f_{\boldsymbol{w}_t}(\boldsymbol{x})
	\label{eq:pred_mean}
\end{equation}

\begin{equation}
	\sigma^2_{\hat{y}} = \mathbb{V}_{p(\boldsymbol{w}|\mathcal{D})}[f_{\boldsymbol{w}}(\boldsymbol{x})] \approx \frac{1}{T}\sum_{t=1}^{T}(f_{\boldsymbol{w}_t}(\boldsymbol{x}) - \hat{y})^2
	\label{eq:pred_var}
\end{equation}

where $f_{\boldsymbol{w}}(\boldsymbol{x})$ represents the network's prediction for input $\boldsymbol{x}$, $\boldsymbol{w}_t$ is the $t$-th weight sample drawn from the approximate posterior distribution, and $T$ is the number of Monte Carlo samples. Through multiple forward passes, we can obtain the mean $\hat{y}$ and variance $\sigma^2_{\hat{y}}$ of the prediction, with the latter being the measure of prediction uncertainty.

\section{Loss Function}

In the proposed Bayesian neural network rPPG model for uncertainty estimation, designing an appropriate loss function is crucial for effectively training the model and accurately capturing the uncertainty in predictions. The loss function should not only guide the network to produce accurate rPPG signal predictions but also promote reliable uncertainty estimation.

\subsection{Combined Loss Function of Pearson Correlation Coefficient and Signal-to-Noise Ratio}

To this end, we adopted a loss function that combines the Pearson correlation coefficient loss and a term based on signal-to-noise ratio (SNR). The loss function $\mathcal{L}$ is defined for a batch of samples as follows:

\begin{equation}
	\mathcal{L} = \frac{1}{N} \sum_{i=1}^{N} \left[ 1 - \rho_i + \lambda \cdot \mathrm{SNR}_i \right],
\end{equation}

where $N$ is the batch size, $\rho_i$ is the Pearson correlation coefficient between the predicted rPPG signal $\hat{\mathbf{y}}_i$ and the true signal $\mathbf{y}_i$ for the $i$-th sample, $\mathrm{SNR}_i$ is the signal-to-noise ratio for the $i$-th sample, and $\lambda$ is a hyperparameter balancing the two terms.

\subsubsection{Pearson Correlation Coefficient}

The Pearson correlation coefficient measures the linear correlation between two signals and is defined as:

\begin{equation}
	\rho_i = \frac{ \sum_{t=1}^{T} \left( \hat{y}_i^{(t)} - \bar{\hat{y}}_i \right) \left( y_i^{(t)} - \bar{y}_i \right) }{ \sqrt{ \sum_{t=1}^{T} \left( \hat{y}_i^{(t)} - \bar{\hat{y}}_i \right)^2 } \sqrt{ \sum_{t=1}^{T} \left( y_i^{(t)} - \bar{y}_i \right)^2 } },
\end{equation}

where $\hat{y}_i^{(t)}$ and $y_i^{(t)}$ are the predicted and true signals for the $i$-th sample at time $t$, and $\bar{\hat{y}}_i$ and $\bar{y}_i$ are their means:

\begin{equation}
	\bar{\hat{y}}_i = \frac{1}{T} \sum_{t=1}^{T} \hat{y}_i^{(t)}, \quad \bar{y}_i = \frac{1}{T} \sum_{t=1}^{T} y_i^{(t)}.
\end{equation}

This term encourages the model to generate predictions that are highly linearly correlated with the true signal, improving signal restoration accuracy.

\subsubsection{Signal-to-Noise Ratio (SNR)}

The SNR term is used to enhance the quality of the predicted signal and is defined as:

\begin{equation}
	\mathrm{SNR}_i = 10 \log_{10} \left( \frac{ P_{\text{signal}} }{ P_{\text{noise}} } \right ) = 10 \log_{10} \left( \frac{ \| \hat{\mathbf{y}}_i \|_2^2 }{ \| \hat{\mathbf{y}}_i - \mathbf{y}_i \|_2^2 } \right ),
\end{equation}

where $P_{\text{signal}}$ represents the power of the predicted signal, $P_{\text{noise}}$ represents the error power between the predicted and true signals, and $\| \cdot \|_2$ denotes the Euclidean norm.

By maximizing SNR, the loss function encourages the model to reduce noise components in the prediction, thereby improving signal quality.

\subsection{KL Divergence for Bayesian Regularization}

In Bayesian neural networks, Kullback-Leibler (KL) divergence is used to regularize the difference between the posterior distribution of network parameters $q(\boldsymbol{\theta})$ and the prior distribution $p(\boldsymbol{\theta})$. KL divergence is defined as:

\begin{equation}
	\mathrm{KL}\left( q(\boldsymbol{\theta}) \| p(\boldsymbol{\theta}) \right) = \int q(\boldsymbol{\theta}) \log \frac{ q(\boldsymbol{\theta}) }{ p(\boldsymbol{\theta}) } d\boldsymbol{\theta}.
\end{equation}

Introducing KL divergence into the loss function helps prevent overfitting and provides a quantification of parameter uncertainty.

\subsection{Final Loss Function}

Combining the above components, the final loss function expression is:

\begin{equation}
	\mathcal{L}_{\text{total}} = \frac{1}{N} \sum_{i=1}^{N} \left[ 1 - \rho_i + \lambda \cdot \mathrm{SNR}_i \right] + \beta \cdot \text{Normalization Factor} \cdot \mathrm{KL}\left( q(\boldsymbol{\theta}) \| p(\boldsymbol{\theta}) \right)
\end{equation}

where $\beta$ is a hyperparameter balancing the data fitting term and Bayesian regularization term in the loss function. By adjusting the values of $\lambda$ and $\beta$, a balance can be achieved between signal accuracy, noise reduction, and uncertainty estimation.

\section{Experiments}

To verify the effectiveness of the method, tests were conducted on the publicly available UBFC-RPPG Dataset \cite{UBFCrppg}, which is a dataset for remote heart rate estimation containing 42 visible light videos from 42 individuals. Each video is approximately 2 minutes long, recorded at 30 frames per second using a Logitech C920 camera with a resolution of 640×480.

To ensure the dataset covers a wider range of heart rate values, all subjects were asked to complete a math game within a specified time to increase their heart rates. The dataset provides physiological indicators such as blood volume pulse and heart rate corresponding to the facial videos.

Following the processing method in \cite{DRNET}, 30 datasets were used for training and the remaining 12 for testing. The BVP signal was normalized to [-1,1], and a fourth-order Butterworth filter with cutoff frequencies [0.6,3] was used for filtering. Cubic spline interpolation was applied to ensure the BVP signal length matched the video length, and videos were randomly flipped. Additionally, we used the AdamW optimizer with a learning rate of $2 \times 10^{-4}$, weight decay of $5 \times 10^{-5}$, a batch size of 4, trained for 50 epochs, and employed a cosine learning rate scheduler.

\subsection{Accuracy Analysis}
For accuracy analysis, we followed the method in \cite{DRbaseline} and used standard deviation (Std), mean absolute error (MAE), root mean square error (RMSE), mean error rate percentage (MER), and Pearson correlation coefficient (r) for performance evaluation. \ref{tab:acc_com}

\begin{table}[htbp]
	\centering
	\caption{Accuracy Testing on the UBFC-RPPG Dataset} 
	\label{tab:acc_com}
	\begin{tabular}{ccccc}
		\toprule
		Method &$HR_{\text{mae}}{\downarrow}$ & $HR_{\text{rmse}}{\downarrow}$ & $HR_{\text{mer}}{\downarrow}$ & $HR_{\text{r}}{\uparrow}$ \\
		
		\midrule
		POS\cite{POS} & 8.35 & 10.00 &9.85\%&0.24\\
		CHROM\cite{chrom} & 8.20 & 9.92 &9.17\%&0.27\\
		GREEN \cite{green}& 6.01 & 7.87 &6.48\%&0.29\\
		SynRhythm\cite{syn} & 5.59 & 6.28 &5.5\%&0.72\\
		PulseGAN\cite{pulse} & 1.19 & 2.10 &1.24\%&0.98\\
		RF-Net(Ours)& 2.56 & 6.60 &2.84\%&0.86\\
		\bottomrule
	\end{tabular}
	\vspace{0.5em}
\end{table}

Although our model has some gap in accuracy compared to the state-of-the-art models, we have the following advantages:
\begin{enumerate}
	\item Simple training without pre-training, no need for additional data, and no need for more complex data processing methods \cite{DRNET}, \cite{pulse}.
	\item Simple model with a small size and fast inference speed.
	\item Capable of uncertainty estimation, making it more practical.
	\item Simple training with fast convergence; MAE can reach below 10 within 5 epochs.
\end{enumerate}

\subsection{Uncertainty Analysis}
\begin{figure}[htbp]
	\centering
	\includegraphics[width=0.8\textwidth]{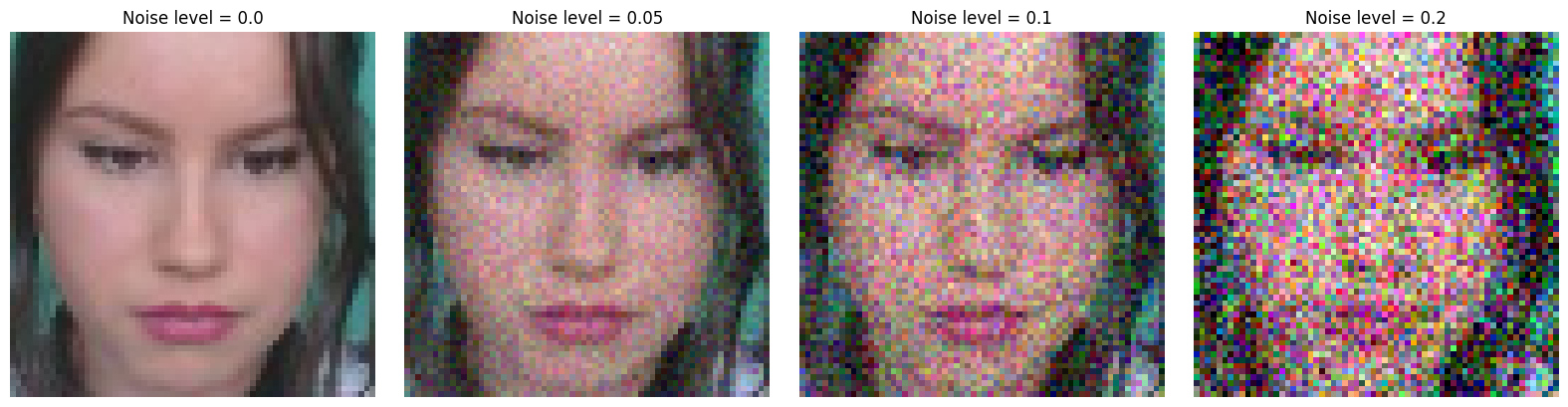}
	\caption{Effects of Different Noise Levels}
	\label{fig:NL}
\end{figure}

Due to the lack of related research on uncertainty analysis in the rPPG field as evaluation criteria, we propose a set of related indicators for uncertainty estimation in the following text. For a more comprehensive analysis of the model's uncertainty estimation capability and robustness, we tested by adding four levels of noise to the input images, as shown in \ref{fig:NL}.

\subsubsection{Correlation Between Uncertainty and Prediction Error}
We used Spearman's rank correlation coefficient to measure the correlation between uncertainty and error.
\begin{enumerate}
	\item \textbf{Uncertainty}
	Following equation \ref{eq:pred_var}, the variance of results obtained from multiple inferences was used as the uncertainty, with the number of inferences uniformly set to 20.
	\item \textbf{Spearman's Rank Correlation Coefficient} 
	Spearman's rank correlation coefficient is a non-parametric statistic used to measure the monotonic relationship between two variables. Unlike the Pearson correlation coefficient, it does not assume a linear relationship between variables or require the variables to follow a normal distribution. Denoted as $\rho$, its value ranges from $[-1, 1]$, and is defined as follows:
	\begin{equation}
		\rho = 1 - \frac{6 \sum_{i=1}^n d_i^2}{n(n^2 - 1)},
	\end{equation}
	where $d_i$ is the rank difference of the $i$-th observation, i.e., the difference in corresponding positions after sorting the two variables, and $n$ is the sample size. Specifically, when $\rho$ approaches $1$, it indicates a strong positive monotonic relationship; when $\rho$ approaches $-1$, it indicates a strong negative monotonic relationship; when $\rho$ approaches $0$, it indicates no monotonic relationship.
	\item \textbf{P-value}
	To evaluate the statistical significance of Spearman's correlation coefficient, we usually compute its corresponding P-value. The P-value indicates the significance of the null hypothesis (i.e., no correlation between the two variables). The smaller the P-value, the more reason to reject the null hypothesis. For example, a significance level of $\alpha = 0.05$ is often chosen; if $P \leq \alpha$, there is significant correlation between the two variables.
	
	In this study, we use Spearman's correlation coefficient $\rho$ to quantify the monotonic relationship between uncertainty estimates (e.g., prediction variance or standard deviation) and prediction errors, and evaluate its statistical significance through P-values \ref{tab:NoiseSP}.
\end{enumerate}

\begin{table}[htbp]
	\centering
	\caption{Relationship Between Noise and Spearman's Rank Correlation Coefficient} 
	\label{tab:NoiseSP}
	\begin{tabular}{ccc}
		\toprule
		Noise Level &Spearman's Rank Correlation Coefficient&P-value \\
		
		\midrule
		0.0 &0.66&0.00 \\
		0.01 &0.63&0.00 \\
		0.05 &-0.0082&0.65 \\
		0.1 &-0.19&0.00 \\
		\bottomrule
	\end{tabular}
	\vspace{0.5em}
\end{table}

The experiments found that uncertainty estimation is effective in no-noise and low-noise conditions, but as noise increases, uncertainty estimation gradually collapses. Due to the lack of related research, no comparative experiments can be conducted here.

\subsection{Relationship Between Noise and Confidence Interval Coverage Rate}
In this study, the distribution of predicted heart rates was obtained through multiple forward passes of the model, allowing calculation of their mean and uncertainty (expressed as variance or standard deviation). By analyzing experimental data under noise conditions, we investigated the impact of noise levels on heart rate prediction and its Confidence Interval Coverage Rate (CICR). A 95\% confidence interval was selected for this experiment, and the proportion of samples where the true value falls within the predicted interval was calculated to evaluate the quality of the model's uncertainty estimation.

To calculate the confidence interval coverage rate, Bayesian neural networks' multiple forward passes were used to sample each sample's prediction mean $\mu$ and variance $\sigma^2$. At a 95\% confidence level, the bounds of the confidence interval are calculated as follows:
\begin{equation}
	\text{Lower Bound} = \mu - z \cdot \sigma, \quad \text{Upper Bound} = \mu + z \cdot \sigma,
\end{equation}
where $z$ is the quantile of the normal distribution, corresponding to a 95\% confidence level when $z = 1.96$. The true heart rate value $\text{HR}_{\text{true}}$ is considered covered by the confidence interval if it meets the following condition:
\begin{equation}
	\text{Lower Bound} \leq \text{HR}_{\text{true}} \leq \text{Upper Bound}.
\end{equation}

The experimental results are shown in \ref{tab:95acc}.

\begin{table}[htbp]
	\centering
	\caption{95\% Confidence Interval Coverage Rate} 
	\label{tab:95acc}
	\begin{tabular}{ccc}
		\toprule
		Noise Level &Coverage Rate &Confidence Interval Width\\
		
		\midrule
		0.0 &0.8698\% &4.5806\\
		0.01 & 0.8593\%&4.78581\\
		0.05 &0.4099 &33.2941\\
		0.1 & 0.3121&25.7304\\
		\bottomrule
	\end{tabular}
	\vspace{0.5em}
\end{table}

The experimental results show that in no-noise and low-noise conditions, the confidence coverage rate is high and does not decrease significantly. However, in medium-high noise conditions, the confidence interval coverage rate drops rapidly. It was also found that in medium and high noise conditions, the confidence interval width is much higher than in no-noise and low-noise conditions, suggesting that confidence interval width can be used as a basis for uncertainty estimation. The overall experimental results can be cross-referenced with the results in \ref{tab:NoiseSP}. Due to the lack of related research, no comparative experiments can be conducted here.

\section{Conclusion}
This paper applies Bayesian Neural Networks (BNN) to rPPG-based physiological signal measurement for the first time, aiming at uncertainty estimation, and proposes a comprehensive benchmark framework for systematically evaluating the performance of different uncertainty estimation methods in the rPPG field. Experimental results indicate that BNNs can provide effective uncertainty estimation under mild to moderate noise conditions; however, as noise levels increase (e.g., under heavy noise conditions), the uncertainty estimation capability of BNNs declines significantly. The study also found that BNNs are sensitive to noise levels, tending to produce wider (or more conservative) uncertainty intervals as noise levels increase. Future research could explore applying Bayesian neural networks to other advanced architectures such as Transformers, Mamba, and more complex data processing methods like STMap for studies on accuracy, robustness, and uncertainty estimation capabilities.

\section{Appendix}
\begin{figure}[htbp]
	\centering
	\includegraphics[width=0.8\textwidth]{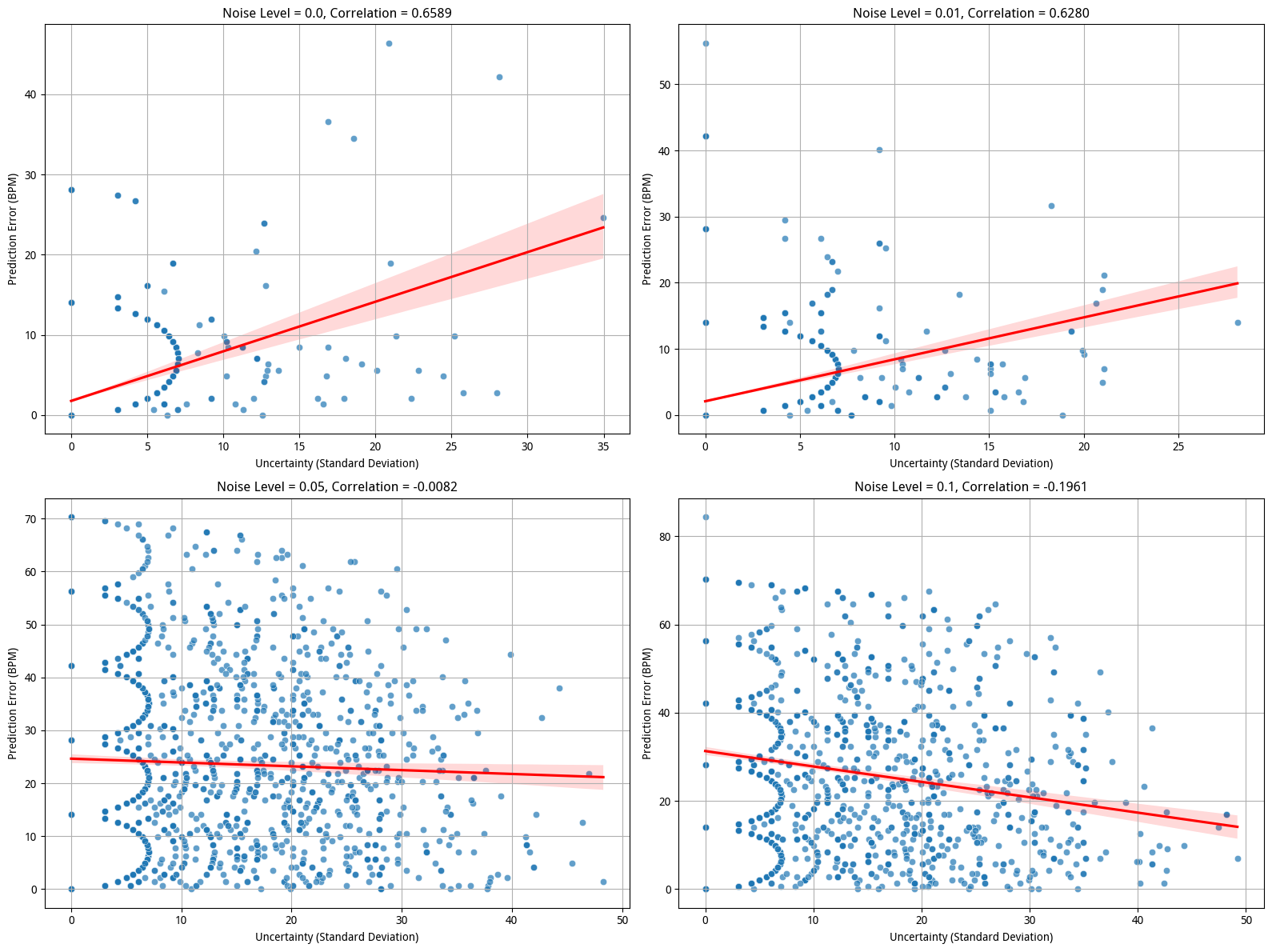}
	\caption{RF-BayesPhysNet Network Structure}
	\label{fig:2UNCandSD}
\end{figure}

During uncertainty estimation, we observed what seems to be a non-linear, periodic relationship between uncertainty and prediction error \ref{fig:2UNCandSD}. We suspect this may be related to the filter experiment or numerical sensitivity in heart rate calculation methods, but it does not affect the conclusions. The exact cause remains to be further analyzed.

\section{Acknowledgments}
Thanks to Qilu University of Technology (Shandong Academy of Sciences) for providing computing resources.

\printbibliography

\end{document}